\documentclass[conference]{IEEEtran}
\usepackage[pdftex]{graphicx}
\usepackage[caption=false]{subfig}
\usepackage{url}
\usepackage{algorithm,algpseudocode}

\newcommand{\cut}[1]{}

\ifCLASSINFOpdf
\else
\fi
\usepackage{amsmath}
\begin{document}
%
\title{Co-training for Demographic Classification Using Deep Learning from Label Proportions}

\author{\IEEEauthorblockN{Ehsan Mohammady Ardehaly}
\IEEEauthorblockA{Department of Computer Science\\
Illinois Institute of Technology\\
Chicago, Illinois 60616\\
Email: emohamm1@hawk.iit.edu}
\and
\IEEEauthorblockN{Aron Culotta}
\IEEEauthorblockA{Department of Computer Science\\
Illinois Institute of Technology\\
Chicago, Illinois 60616\\
Email: aculotta@iit.edu}
}


%


\maketitle

\begin{abstract}
Deep learning algorithms have recently produced state-of-the-art accuracy in many classification tasks, but this success is typically dependent on access to many annotated training examples. For domains without such data, an attractive alternative is to train models with light, or distant supervision. In this paper, we introduce a deep neural network for the Learning from Label Proportion (LLP) setting, in which the training data consist of bags of unlabeled instances with associated label distributions for each bag. We introduce a new regularization layer, {\bf Batch Averager}, that can be appended to the last layer of any deep neural network to convert it from supervised learning to LLP. This layer can be implemented readily with existing deep learning packages. To further support domains in which the data consist of two conditionally independent feature views (e.g. image and text), we propose a co-training algorithm that iteratively generates pseudo bags and refits the deep LLP model to improve classification accuracy. We demonstrate our models on demographic attribute classification (gender and race/ethnicity), which has many applications in social media analysis, public health, and marketing. We conduct experiments to predict demographics of Twitter users based on their tweets and profile image, without requiring any user-level annotations for training. We find that the deep LLP approach outperforms baselines for both text and image features separately. Additionally, we find that co-training algorithm improves image and text classification by 4\% and 8\% absolute F1, respectively. Finally, an ensemble of text and image classifiers further improves the absolute F1 measure by 4\% on average.
\end{abstract}


%
\IEEEpeerreviewmaketitle

\section{Introduction}

Deep learning methods have produced state-of-the-art accuracy on many different classification tasks especially for image classification\cite{DBLP:journals/corr/SimonyanZ14a,DBLP:journals/corr/HeZRS15,DBLP:journals/corr/Chollet16a,Going:2015:Szegedy,DBLP:journals/corr/SzegedyIV16}. Because these networks typically have millions of parameters, they rely on access to large labeled data sets such as ImageNet~\cite{ILSVRC15}, which has over one million annotated images. While transfer learning can help adapt a network to a new domain~\cite{DBLP:journals/corr/YosinskiCBL14}, it still requires labeled data on the target domain.

An attractive alternative is Learning from Label Proportion (LLP), in which the training samples are divided into a set of bags, and only the label distribution of each bag is known. The main advantage of LLP is that it does not require annotations for individual instances. Furthermore, in many domains label proportions are readily available --- for example, by associating geolocated social media messages with county population statistics, we can fit a model of demographics without annotating individual users.

While many LLP models have been proposed based on the logistic hypothesis \cite{mann2007simple}, SVM \cite{Lai:2014:VED}, and graphical models \cite{gracca2007expectation}, there has been little work that considers deep learning. In this paper, we propose an approach which converts a deep neural network from a supervised classifier to LLP. This method can readily be implemented in popular deep learning packages by introducing a new regularization layer into the network. We propose such a layer, called the {\bf Batch Averager}. Similar to label regularization~\cite{mann2007simple}, this layer computes the average of its input as the output. Like other regularization layers, this layer is only applied at training time. The {\bf Batch Averager} is typically appended to the last layer of a network to convert it from a supervised learning to a deep LLP. We use KL-Divergence as the error function to train the network to produce predictions that match the provided label proportions.

This deep LLP framework has these key advantages:
\begin{itemize}
\item {\bf Simplicity}: The {\bf Batch Averager} is based on a very simple tensor operation (average) and, unlike the {\bf Batch Normalizer} layer~\cite{DBLP:journals/corr/IoffeS15}, it does not have any training weights and does not noticeably affect training time.
\item {\bf Compatibility}: The framework is fully compatible with almost every deep learning packages, and requires only a few line of codes to implement. We have tested the approach with both Theano~\cite{DBLP:journals/corr/Al-RfouAAa16} and Tensorflow~\cite{DBLP:journals/corr/AbadiABBCCCDDDG16}.
\item {\bf Availability}: The framework can convert almost every supervised classification network to LLP. 
\item {\bf Accuracy}: Our empirical results indicate that the deep LLP has a comparable accuracy to supervised models subject to proper constraints to generate bags.
\end{itemize}

In addition, we also propose co-training with deep LLP to further improve accuracy. In some applications, multiple views of the data are available (e.g. text and image). For example, many social media sites contain both image and textual data. While there have been many deep learning methods that directly combine text and image features (e.g., \cite{Sakaki:2010:EST:1772690.1772777}), we instead require a model that is robust to cases where one feature view is missing. For example, some users may not post images, yet we would still like to classify based on the text. An attractive alternative is co-training~\cite{Blum:1998:CLU:279943.279962}. Because traditional co-training requires labeled data, we propose a new algorithm that is more suitable for the LLP setting. In this method, we use an image-based LLP model to create bags with estimated label proportions, which we call {\it pseudo-bags}. We then fit a text-based LLP model on these pseudo-bags, and use it to in turn create pseudo-bags for the image-based model.

\cut{ The algorithm proceeds iteratively to improve the accuracy of each model. Our experimental results show that this algorithm improves both text and image classification tasks without using any labeled data. We also find that applying an ensemble method after training can further improve accuracy.}

We conduct experiments classifying Twitter users into demographic categories (gender and race/ethnicity) based on the profile image and tweets. We find that the proposed deep LLP framework outperforms both supervised and LLP baselines, and that co-training further improves accuracy of the models.

The remainder of the paper is organized as follows. In Section~\ref{sec.related}, we review related work on LLP and co-training, and Section~\ref{sec.models} provides our deep LLP framework and co-training algorithm. In Section~\ref{sec.data} we describe the data collected for the experiments.  In Section~\ref{sec.results} we present our empirical results; Section~\ref{sec.conclusions} concludes and describes plans for future work.

\section{Related Work}
\label{sec.related}

\cut{While support vector machines often work better with a small amount of labeled data without outliers \cite{Forman2004},}
Deep learning methods have produced state-of-the-art results on many classification tasks, but typically require many labeled training instances. For image classification, researchers usually compare their model by training on large datasets such as ImageNet \cite{ILSVRC15}. VGG-16 network (2014) achieves 90.1\% top-5 accuracy (single crop) \cite{DBLP:journals/corr/SimonyanZ14a} on ImageNet by stacking 16 layers on the top of each other. ResNet-152 network (2015) improves that to 93.3\% by introducing the residual networks \cite{DBLP:journals/corr/HeZRS15}. Inception V1-V3 networks present inception module for convolutional neural network \cite{Going:2015:Szegedy,DBLP:journals/corr/IoffeS15,DBLP:journals/corr/SzegedyVISW15}, and Inception-V3 (2015) achieves 94.1\% top-5 accuracy on ImageNet. The inception module can be combined with the residual networks to create Inception-ResNet-v2 (2016) with 95.1 \% accuracy \cite{DBLP:journals/corr/SzegedyIV16}. XCeption (2016) network introduces extreme inception module by using deepwise separable convolution layers \cite{DBLP:journals/corr/SifreM14} and achieves 94.5\% top-5 accuracy \cite{DBLP:journals/corr/Chollet16a}.

All of these models highly rely on the vast amount of labeled data. However, with transfer learning, the pre-trained weights (using ImageNet) can move to other image classification tasks \cite{DBLP:journals/corr/YosinskiCBL14}. But, still we need annotated data on the target domain, and alternative approaches such as LLP are required in the absence of labeled data.

Only a handful of researchers attempt to use  deep learning for LLP settings. Kotzias et al.~\cite{Kotzias:2015:GIL:2783258.2783380} propose a model for the particular case of LLP when the label of bags are available. For example, for text classification task, when we have the label of a bag of multiple sentences, they propose a model to infer the label of each sentence using an objective function to smooth the posterior probability of samples based on sample similarity and bag constraints. They use a convolutional neural network to infer sentence similarity for text classification. 

Other researchers provide models for image segmentation. In this setting, each image is split into multiple small images (as a bag), and the classifier tries to infer labels of these regions using the label proportion of the bag. Li et al.~\cite{li2015alter} suggest a convolutional neural network with a probabilistic method to estimate labels by considering the proportion bias, using the Expectation Maximization algorithm to determine model parameters. In their approach, they use satellite images with know ice area ratio to train a classifier to predict label of small segments of images.

While these methods have promising results on a particular domain, to the best of our knowledge, no method has proposed a framework that can be readily applied to diverse classification tasks. Inspired by label regularization \cite{mann2007simple}, we fill this gap by introducing {\bf Batch Averager} as a regularizer layer.

The traditional L2 regularization appears to not be sufficient for deep neural network because overfitting is a severe problem in these networks. Srivastava et al.~\cite{Srivastava:2014:DSW:2627435.2670313} introduce a Dropout layer that randomly drops some units in backpropagation step and show that it can significantly reduce overfitting. Furthermore, the Batch Normalizer is introduced to normalize the output of cells and reduce internal covariate shift of weights~\cite{DBLP:journals/corr/IoffeS15}. Similar to these two normalization layers, our proposed {\bf Batch Averager} layer applies only at training time (Batch Normalizer uses the moving average and standard deviation to normalize the output at  testing time without updating the moving average and standard deviation).

Recently, demographic classification with deep learning has been proposed by researchers. Zhang et al.~ \cite{DBLP:journals/corr/ZhangPRDB13} offer a method to infer demographic attributes (gender) from the wild (unconstrained images). Similarly, Liu et al.~\cite{DBLP:journals/corr/LiuLWT14} propose convolutional neural network to classify attributes such as age, gender, and race in the wild. Ranjan et al.~\cite{DBLP:journals/corr/RanjanPC16} provide a fast RCNN to localize and recognize all faces in the scene with their attributes (e.g. gender, pose). Few attempts to use both textual and image features together.

Co-training is a semi-supervised method that trains on two views of features on a small set of labeled data and iteratively adds pseudo labels from a large set of unlabeled data \cite{Blum:1998:CLU:279943.279962}. Gupta et al.~\cite{Gupta:2008:WLL:1431932.1431984} demonstrate a co-training algorithm that trains on captioned image to recognize human actions with SVMs. Their model also learns from videos of athletic events with commentary. 

The majority of these works use captions to leverage image classification accuracy by co-training methods, and cannot be applied to text only classification purpose. Furthermore, they require annotated labeled data. On the other hand, our proposed co-training model trains both image and text classifiers that can be used separately. Additionally, we do not need any labeled data; and if the testing sample has both image and text features, we can apply an ensemble model (by soft voting) to improve the classification accuracy.

\section{Models}
\label{sec.models}

In this section, we first define our proposed regularization layer, {\bf Batch Averager}. Then, we illustrate the deep LLP framework with this layer. Next, we provide an algorithm to create bags with appropriate label distributions for training. Finally, we offer a co-training approach for training on two views of data (e.g. text and image) in LLP settings.

\subsection{Batch Averager Layer}

In the Learning from Label Proportion setting, the training data are divided into bags, and only the label distribution for each bag is known. For bag $i$, let $B_i$ be the set of samples in this bag; i.e. $B_i = \{X_{i,j}\}$ where $X_{ij}$ is the feature vector for instance $j$ in bag $i$. Let $\tilde{y_i}$ be the provided label proportion for bag $i$, and $T_i$ be the number of samples in this bag (i.e. $T_i=|B_i|$). E.g., for binary classification, $\tilde{y_i}$ is the fraction of positive instances in bag $i$.

To implement Deep LLP, we assume that each bag is assigned to exactly one batch for gradient optimization. Also, because deep neural networks are typically trained by GPU cores with limited memory, there is usually a maximum batch size determined by the number of parameters that can fit into GPU memory. As a result, if a bag is too large, we need to break it down into smaller bags.

Our work is inspired by label regularization \cite{mann2007simple}, and we define a regularization layer for a neural network. The label regularization model estimates bag posterior probabilities by the average of the posterior probabilities of the instances in that bag. In a neural network, this can just be implemented by computing the average of the output of the last layer per batch. As a result, we name this layer {\bf Batch Averager}. 

Let $y$ be the (unobserved) output of the last layer of a supervised network. Because in classification tasks, the last layer is typically {\bf logit} function (either softmax or logistic), it returns a vector of posterior label probabilities; i.e. for bag $i$, instance $j$, and class $k$ we have:

\begin{equation}
  P(y=k|X_{i,j}) = y_{i,j}^{(k)}
\end{equation}

Inspired by label regularization, we estimate the posterior probability of bag $i$, $\bar{y_i}$, as the average of posterior probability of all instances in bag $i$; i.e.
\begin{equation}
  \bar{y_i}^{(k)} = \frac{1}{T_i} \sum_{j \in B_i} P(y=k|X_{i,j}) = \frac{1}{T_i} \sum_{j \in B_i} y_{i,j}^{(k)}
  \label{eq:posterior}
\end{equation}

We expect that the bag posterior probability ($\bar{y_i}$) should be close to the bag prior probability ($\tilde{y_i}$). Again, similar to label regularization, we use KL-divergence as the error function between posterior and prior, and the target of the neural network is to minimize this error function:

\begin{equation}
 \Delta(\tilde{y}, \bar{y}) = \sum_k \tilde{y}^{(k)} \log \frac{\tilde{y}^{(k)}}{\bar{y}^{(k)}}
\end{equation}

The {\bf Batch Averager} layer is very similar to the Batch Normalizer layer~\cite{DBLP:journals/corr/IoffeS15}. The latter normalizes the output of the batch to have zero mean and unit variance; the former converts the input layer to its average. As a regularization layer, {\bf Batch Averager} is only applied at training time, not at testing time. Typically, {\bf Batch Averager} would be applied to the last layer of the network, immediately after the Logit layer.

Implementing the {\bf Batch Averager} is straightforward with most current open-source deep learning packages. However, most of these packages assume that the output of the network is a tensor with the same size as batch size ($T_i$). As a result, we need to repeat the average for $T_i$ times. Similarly, we need to repeat the prior ($\tilde{y_i}$) for $T_i$ times. Because the number of samples per bag (batch size) is typically different per bag, we need to add sample weight $\frac{1}{T_i}$ for each sample in bag $i$. As a result, the entire bag has sample weight one.

More formally, for bag $i$, we train the network in a  similar fashion as traditional supervised learning with feature vectors $[X_{i,1}, ... X_{i,T_i}]$, labels $[\tilde{y_i}, ...,\tilde{y_i}]$, and sample weights $[\frac{1}{T_i}, ...,\frac{1}{T_i}]$. We additionally use KL-divergence as the error function for back propagation.

We use Keras\footnote{http://www.keras.io} to implement this layer, and it supports both the Tensorflow \cite{DBLP:journals/corr/AbadiABBCCCDDDG16} and Theano \cite{DBLP:journals/corr/Al-RfouAAa16} backend. The implementation is very simple (just one line in Keras), and can extend to other deep learning packages. Another implementation aspect to take into consideration is that by default deep learning packages assume fixed batch size. However, the bag size in LLP is typically dynamic. To support dynamic batch size, we implement a generator code to make batches and fit in the network. Also, this generator code automatically breaks down large bags to randomly smaller bags that can fit in the GPU memory.

To implement this layer with dynamic batch size, we use the broadcasting feature in tensor operations. Suppose $K$ is the backend (Tensorflow or Theano), we define this function as the activation function for {\bf Batch Averager} layer:
\begin{equation}
 x \times 0 + K.mean(x, 0)
\end{equation}
where $x$ is the input tensor of the layer. Because the {\bf Batch Averager} is appended after the {\bf logit} layer, $x$ is a two-dimensional tensor with size $T_i \times N$, where $N$ is the number of output classes. This function, first creates a zero tensor with the same size of $x$. Then it computes the average of $x$ over the first axis as a vector tensor with size $N$. Finally, it adds them together by using the broadcasting feature, and creates a tensor with the same size of $x$, that the average of $x$ is repeated over the first axis for $T_i$ times.

\subsection{Deep LLP framework}

In this section, we provide a framework to implement deep LLP networks. This framework uses the {\bf Batch Average} regularizer as the last layer, and can be applied in any classification application (e.g. text, image). We furthermore propose text classification and image classification networks.

Similar to any classification deep network, in Deep LLP framework, the input of the network is the feature vectors (i.e. textual or image feature). Then we feed the network with layers that are typically used for the classification task at hand. The next layer is the Logit (softmax) layer to compute class probabilities. Finally, we add {\bf Batch Averager} for label regularization, which sends the average of features to the output.

\cut{
\begin{figure}[t]
		\centering \includegraphics[scale=0.1,angle =90]{plots/framework.png}
    \caption{The generic deep LLP framework; the supervised network is the deep or shallow network for classification task.}
    \label{fig.framework}
\end{figure}
}

\begin{figure*}[t]
    \centering \includegraphics[scale=0.6]{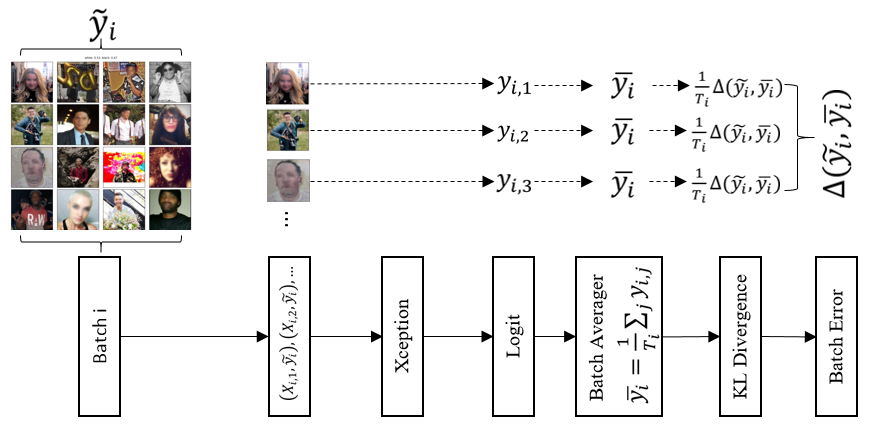}
    \caption{The deep LLP framework applied to image classification.}
    \label{fig.framework2}
\end{figure*}

For text classification, we simply use a shallow network with only one dense layer (with 16 cells), followed by a dropout layer to avoid overfitting. We also use a temperature parameter as suggested in previous work \cite{mann2007simple}. The temperature can be readily implemented in the Logit layer by a tensor operation.

For image classification, we use  Xception, a state-of-the-art image classification model~\cite{DBLP:journals/corr/Chollet16a}. We use the pre-trained weights (trained on ImageNet), and freeze the first two blocks of the network to train it with the maximum batch size 32 and fit the network with color images with dimension $299\times299$. Figure~\ref{fig.framework2} shows a more detailed network for this model.

\subsection{Bag creation algorithm}

In some domains, bags are naturally available with soft constraints such as geolocation, allowing us to assign label proportions to groups of instances. For example, in experiments below, we attach the U.S. Census statistics of county demographics to Twitter users from the same location. However, in other domains, our constraint is a prior probability based on an attribute of an instance. For example, according to US Census, 67\% of people with the last name `Taylor' are white. The problem is to construct a bag that combines many of these constraints and associate an accurate label proportion for training.

For class $y$ and unlabeled samples $\{X_i\}$, we assume we have a prior probability for this class $P(y|X_i)$ (e.g., $P(\mathrm{white} \mid \mathrm{Taylor}) = .67)$. The goal is to select instances from $X_i$ to construct a bag, then to assign the expected proportion of that bag that have class $y$. Algorithm 1 shows how to create bags for this class. The algorithm takes as input a maximum bag size ($N$) and a threshold ($t$). In the results below, we set $N$ to 64 for all experiments. The algorithm first removes samples with probability lower than the threshold $t$. The threshold is typically greater than .5 (in preliminary experiments, the results were not very sensitive to this parameter and we tune this threshold for each task; e.g. .8 for gender classification and .9 for race classification). Then, samples are sorted by decreasing order of their prior probability. Next, we move first $N$ sorted samples to the first bag, and the next $N$ samples to the next bag and continue until all remaining samples (above the threshold) are assigned to bags. Finally, we compute the bag label proportion as the average of the prior probability of samples inside the bag. These bags are then used as training data in the deep LLP model.

The advantage of this bag creation approach is that it allows us to associate label proportions with instances using pre-existing population statistics. Of course, these label proportions are likely to be inexact, and contain some selection bias due because Twitter users are not representative of the population. However, several prior works have found that LLP methods are robust to this type of noise~\cite{mann2007simple}; we also find this to be the case in the present work.

\begin{algorithm} [t]
\caption{: This algorithm create bags with class prior and returns them.
{\bf Parameters}: Prior probability $P$, class $y$, feature vectors $X_i$, threshold $t$, and maximum batch size $N$.}
\label{alg:bag}
\begin{algorithmic}[1]
\Procedure{CreateBags}{$P, X_i, y, t, N$}
	\State $U \gets \{X_i : P(y|X_i) > t\}$
	\State $U \gets$ Sorted(U, key = $P(y|X_i)$) \Comment Descending order.
	\State $B \gets []$
	\State $\tilde{y} \gets []$
	\While{$U \neq \emptyset$}
	  \State $S \gets$ first $N$ samples in $U$
		\State $B$.append($S$)
		\State $\tilde{y}$.append(Average($\{P(y|x): \forall x \in S\}$))
		\State $U \gets U \setminus S$
	\EndWhile
	\State return $B$, $\tilde{y}$
\EndProcedure
\end{algorithmic}
\end{algorithm}

\subsection{Co-training for LLP}

In this section, we provide an algorithm for applying co-training in the LLP setting. This algorithm is useful when we have two conditionally independent feature views of data. An interesting case is when we have both text and image views of the data. Some views may naturally have more bags than others, or may have more accurate label proportions associated with them. Thus, our goal is to combine the advantages of each view to produce a more accurate classifier.

To apply our algorithm, we assume we have two sets of bags, one with textual features and one with image features. 
\cut{Also, we need a set of unlabeled data with both textual and pictures. Unlike traditional co-training, because we do not have labeled data, the unlabeled samples can overlap with training data. However, we expect the algorithm works better with completely separate unlabeled data.}
Let $B_k$, $\tilde{y}_k$, $U_k$ be bags, label proportions, and unlabeled data for view $k$ (e.g. text or image), where $U_k$ refer to the same set of instances. We propose Algorithm~\ref{alg:cot} for co-training in LLP settings. 
The algorithm proceeds by using the model trained on one view to create {\it pseudo bags} for the other view. The algorithm is initialized with empty pseudo bags ($B_0, \tilde{y}_0$). For each iteration, we first train a deep LLP model on the union of the current view (e.g. text or image) and the pseudo bags for that view. \cut{Then, it reports evaluation results.} Next, it predicts the posterior probability of unlabeled samples with the current view. Then, it calls Algorithm~\ref{alg:bag} to create pseudo bags, using the current view's posteriors as the priors $P(y|x)$. Finally, it switches views for the next iteration. In the experiments below, this algorithm tends to converge quickly (e.g., six iterations).

\begin{algorithm} [t]
\caption{: The co-training algorithm for LLP settings.
{\bf Parameters}: Bags $B^{(k)}$, label proportion $\tilde{y}^{(k)}$, and unlabeled features $U^{(k)}$ for both views.}
\label{alg:cot}
\begin{algorithmic}[1]
\Procedure{CoTrainingLLP}{$(B_1, \tilde{y}_1, U_1), (B_2, \tilde{y}_2, U_2)$}
	\State $B_0 \gets \emptyset$
	\State $\tilde{y}_0 \gets \emptyset$
	\While{Stopping condition}
	  \State dllp.train($B_1 \cup B_0, \tilde{y}_1 \cup \tilde{y}_0$) \Comment Training step.
		\State P $\gets$ dllp.predict($U_1$) \Comment Create posterior.
		\State $B_0, \tilde{y}_0$ $\gets$ CreateBags($P, U_1$) \Comment Create bags.
		\State $(B_1, \tilde{y}_1, U_1) \leftrightarrow (B_2, \tilde{y}_2, U_2)$ \Comment Switch views.
	\EndWhile
\EndProcedure
\end{algorithmic}
\end{algorithm}

\section{Data}
\label{sec.data}

To evaluate our deep LLP framework, we consider the task of predicting the gender and race/ethnicity of a Twitter user based on their tweets and profile image. In this section, we first describe how we collect data from Twitter and create bags, as well as the annotated data used for validation.\footnote{Replication code and data will be made available upon publication.}

\subsection{Twitter data}

It is common that image data comes with metadata. This metadata is useful to create bags. For example, images in Flickr\footnote{http://www.flickr.com} have caption and description. Also, Twitter users often have a profile image. To demonstrate how the metadata can be used as a constraint to create bags, we collect data from Twitter. For the purpose of this study, we use geolocation and name in the metadata as constraints to create bags.

First, we use the Twitter Streaming API to collect roughly 120K tweets and remove users without an image profile; approximately 33K tweets with an image profile remain. However, not all of users use their own photo. To decrease noise, we want to ensure that there is only one face in the image profile. To do so, we apply the Viola-Jones object detection algorithm~\cite{DBLP:conf/cvpr/ViolaJ01}. This algorithm detects multiple faces in a scene with a low false positive rate. Finally, roughly 10.5K images with exactly one identified face remain. (Please note that this filtering only affects training data; validation data may contain multiple faces.) For each user, we download the most recent 200 tweets for use by the text-based model. 

\begin{figure}[t]
		\centering \includegraphics[scale=0.16]{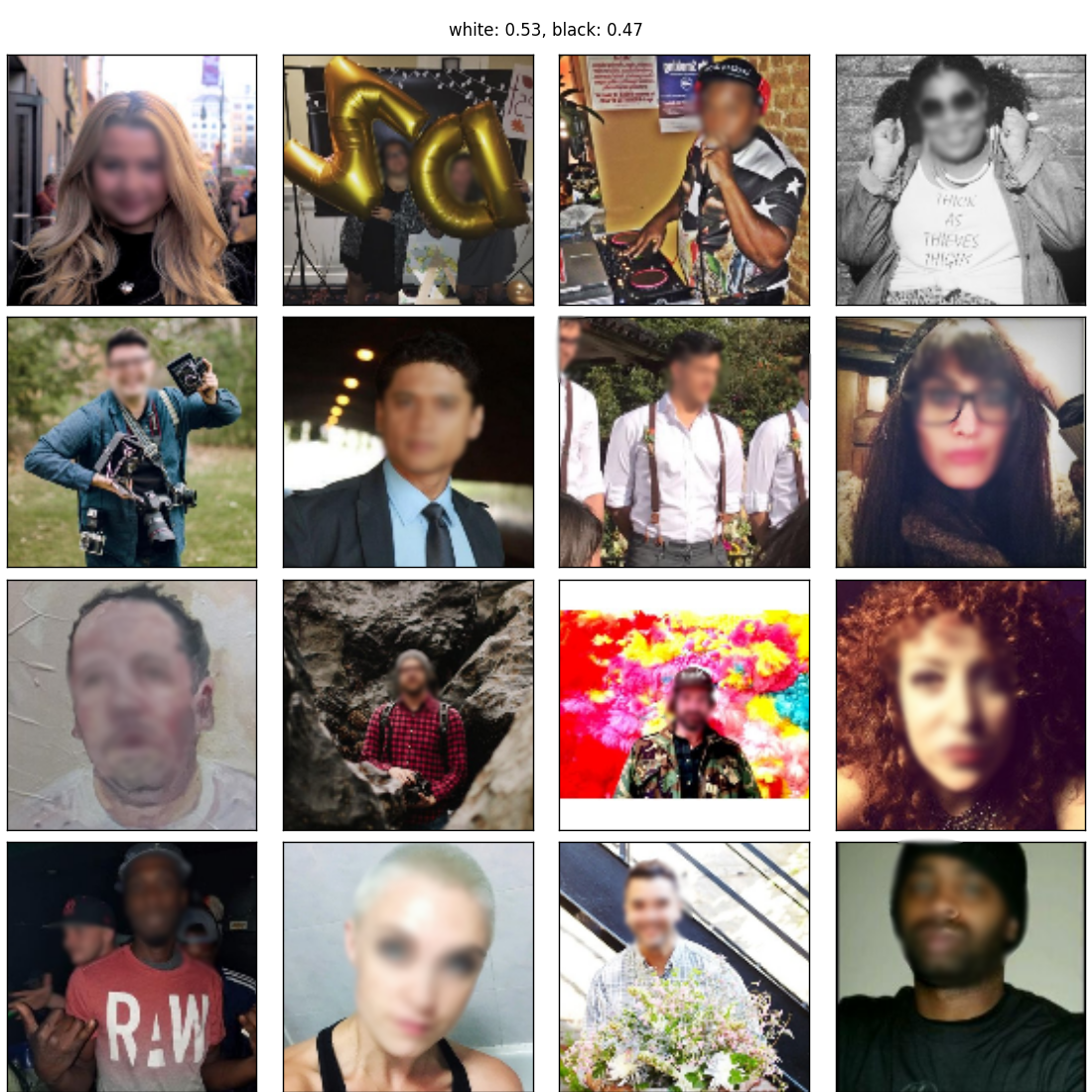}
    \caption{A county bag for race classification with 53\% White American prior probability.}
    \label{fig.race}
\end{figure}

Next, we use the county and name metadata to create bags. We use the county constraint as described in recent work~\cite{ehsan2014using} by associating each user with a county in the U.S., based on the geolocation information provided in their tweets. This results in 85 bags with an average  of 124 users per bags. Figure~\ref{fig.race} illustrate a generated bag using the county constraint with 53\% prior probability of class "white." This figure shows that there are some photos with multiple faces or cartoon faces, due to errors in the face detection algorithm. So, we expect that our photo bags will contain some noise.

\begin{figure}[t]
		\centering \includegraphics[scale=0.16]{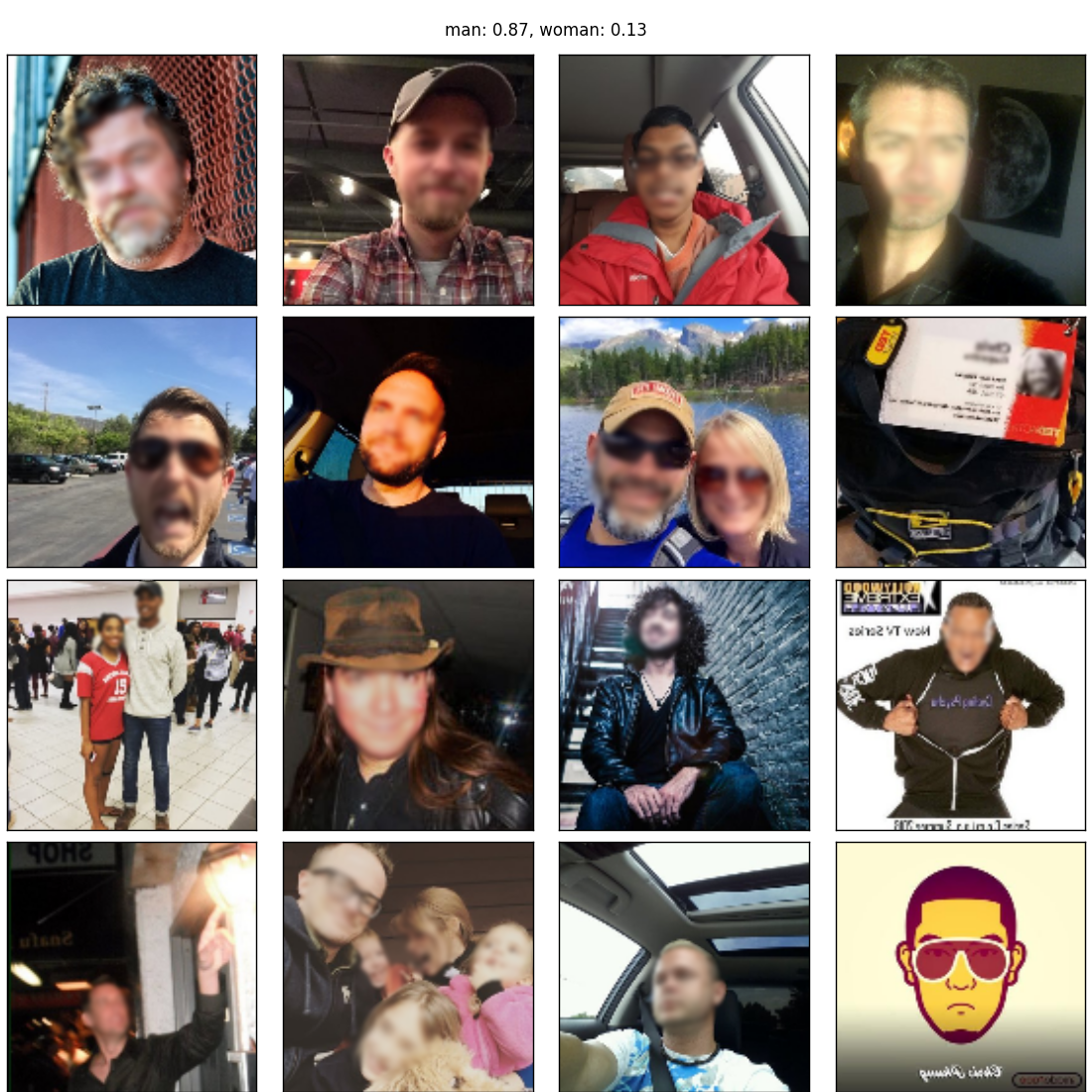}
    \caption{A gender bag that is created by name constraints with 87\% male prior probability.}
    \label{fig.gender}
\end{figure}

We also use name attribute (where available in the user's profile) to estimate label proportions. For gender classification, we use the first name with the data from US Social Security Administration (SSA). For example, according to SSA baby data\footnote{https://www.ssa.gov/oact/babynames/}, for the first name `Casey', the probability of being a man is 59\%. Then we use Algorithm~\ref{alg:bag} with $N=64$ and $t=.6$ to create bags. Figure~\ref{fig.gender} shows a bag using name constraints with 87\% male prior probability. There is again some noise due to face detection. Additionally, for race/ethnicity classification, we use the last name as described in recent research \cite{ehsan2014using} to estimate class priors for users who provide their last name, and run Algorithm~\ref{alg:bag} to create bags. For example, according to US Census, 67\% of people with the last name `Taylor' are white.

Finally, for evaluation purposes, we manually annotate 320 photos; the class distribution is shown in Table~\ref{tab:dist}\cut{some of the pictures that the Viola-Jones algorithm fails to find one face, to one of the classes. We totally annotate 320 photos,}. We use this dataset only for the evaluation purposes, not for training.

\begin{table}[t]
\centering
\caption{The distribution of evaluation set (manually annotated):}
\label{tab:dist}
\begin{tabular}{lcc}
Race     & Male & Female \\ \hline
Hispanic & 15   & 15     \\
White    & 105  & 105    \\
Black    & 40   & 40     \\ \hline
\end{tabular}
\end{table}

\begin{table}[t]
\centering
\caption{The distribution of CFD dataset:}
\label{tab:cfd}
\begin{tabular}{lcc}
Race     & Male & Female \\ \hline
Hispanic & 52   & 56     \\
White    & 228  & 236    \\
Black    & 231  & 295    \\ \hline
\end{tabular}
\end{table}

\subsection{Google search}

Figure~\ref{fig.race} and Figure~\ref{fig.gender} show that Twitter profile images are often noisy. Furthermore, the class distributions are unbalanced, since African-Americans are less frequent in both county and name constraints. This motivates a third type of constraint for image classification. We submit keywords (e.g. `Latino woman' and 'Black American man') to Google images search to identify images that are likely to come from the desired class. Then, we apply the Viola-Jones algorithm to remove photos without exactly one face in the scene. Because Google search sort images in decreasing order of relevance, we associate decreasing label proportions as we descend the list. Specifically, we group search results into bags of size 64 for each contiguous set of results, up to a maximum of 800 results. The first bag is assigned a label proportion of 95\% for the positive class, and the proportion is reduced by 5\% for each subsequent bag, to a minimum of 55\%. 

\subsection{CFD dataset}

For additional validation, we use the Chicago Face Database (CFD), which contains high-quality frontal images (without background) for both genders and four racial/ethnic categories (white, black, Hispanic, and Asian), and facial expressions \cite{Ma2015}. We remove Asian photos (because only few Asian samples are in our Twitter data) and use this dataset as an additional testing set. The advantage of this database is that it has more samples for the Hispanic category than our validation set. Table~\ref{tab:cfd} shows the class distribution of the CFD database.

However, this dataset has a different distribution of our training set. Our training set mostly has a wild condition (not frontal, with background and lower quality). As a result, we expect a different behavior of the classifier for this set. As we will discuss below in the experimental results, because these images have a very high quality, our race/ethnicity classifier is extremely accurate (with 98\% F1 measure for black and white class). However, the F1 metric drops for gender classification because the lack of body features for CFD dataset.

\section{Results}
\label{sec.results}

For the purpose of this study, we train image and textual classifiers for the demographic attribute (gender and race/ethnicity) task. For race classification, we consider three categories (white, black, and Hispanic). However, according to the US Census\footnote{http://www.census.gov/prod/cen2010/briefs/c2010br-04.pdf}, Hispanic is a debated term that refers to Spanish culture or origin regardless of race and can be of any ethnicity. As a result, we consider another classification task restricted only to black and white classes. We refer to the latter as {\bf `race2'} and the former as {\bf `race3'}  classification task. Since labeled samples for both Twitter and CFD datasets are imbalanced, we report the weighted average F1 measure to compare results.

In this section, we first provide experimental results for text classification task and compare it with the state of the art baselines. Then, we describe image classification results and compare them with third-party face APIs. Finally, we demonstrate how co-training improves both text and image classification, as well as the advantage of an ensemble approach.

\subsection{Text classification result}

Because the search constraints created by searching for Google images do not have any textual features, for text classification task we only use county and name constraints. Table~\ref{tab:text:const} shows the weighted F1 measure for different tasks with all combination of constraints. According to this table, the county constraint results in higher accuracy for race classification, and name bags have higher accuracy for gender classification. That means that location is more informative for the race classification, and first name is more informative for the gender classification. Also, for the race task, using both constraints together increases the classification accuracy, but for the gender classification, the result of combining constraints is almost same as using only name bags.

Since using both constraints improves (or at least does not harm) the classification task, we use both name and county constraints in all subsequent text classification experiments. We use the maximum batch size of 32 to generate Table~\ref{tab:text:const}; to do so, any larger bags are randomly split into smaller bags with the maximum size of 32 at each iteration of training.

\begin{table}[t]
\centering
\caption{The weighted average F1 measure for different tasks and constraints for text classification on Twitter dataset.}
\label{tab:text:const}
\begin{tabular}{lcccc}
Model       & {\bf race2} & {\bf race3} & {\bf gender} & Average \\ \hline
county      & 78    	 		& 64    			& 54     		 	 & 65			 \\
name        & 61       		& 55    			& {\bf 83 } 	 & 66			 \\
county-name & {\bf 80} 		& {\bf 73} 	  & {\bf 83 }    & {\bf 79}\\ \hline
\end{tabular}
\end{table}

Table~\ref{tab:text:batch} presents the weighted average F1 metric for different maximum batch sizes using county and name constraints. According to this table, while gender classification is stable for different batch sizes, the maximum batch size of 32 works better for race tasks. This result is not surprising, because a batch size of 16 is too small for a bag, and batch size of 32 has the advantage of randomly splitting bags (that is created with size 64) to avoid overfitting. An interesting case is batch size of 1, which associates a label distribution to individual instances. The problem with this approach is that it is less robust to noise --- for example, if a label proportion of 90\% positive is applied to a negative instance, the classifier will attempt to fit this noisy example. However, if this instance is the only negative one in a batch of size 10, then the classifier is given the flexibility of predicting it as negative while still optimizing the loss function. Indeed, we find that batch size 1 is not effective on this task.

\begin{table}[t]
\centering
\caption{The weighted average F1 measure with different batch size for text classification on Twitter dataset.}
\label{tab:text:batch}
\begin{tabular}{ccccc}
Max batch size & {\bf race2} & {\bf race3} & {\bf gender} & Average \\ \hline
16			    	 & {\bf 80} 	 & 71    			 & {\bf 83 } 	  & 78			\\
32 						 & {\bf 80} 	 & {\bf 73} 	 & {\bf 83 } 	  & {\bf 79}\\
64		      	 & 73    	 		 & 64    			 & {\bf 83 } 	  & 73			\\ \hline
\end{tabular}
\end{table}

Finally, we compare our deep LLP model with state-of-the-art shallow LLP models --- we use ridge regression for LLP~\cite{ehsan2014using} and label regularization~\cite{mann2007simple} for comparison. Table~\ref{tab:text:bl} compares our model with maximum batch size 32 with other models for county and name constraints. According to this table, while our model has  the same F1 as ridge LLP for {\bf race2} task, it has higher F1 for all other tasks. More specifically, on average, while both ridge LLP and label regularization have the same F1 score (76), our model has the highest F1 measure (79).

\begin{table}[t]
\centering
\caption{Comparison with the state of the art LLP text classification baselines on Twitter dataset.}
\label{tab:text:bl}
\begin{tabular}{lcccc}
Model Name				   & {\bf race2} & {\bf race3} & {\bf gender} & Average  \\ \hline
Deep LLP			 			 & {\bf 80} 	 & {\bf 73} 	 & {\bf 83}   	& {\bf 79} \\
Ridge LLP 					 & {\bf 80}	 	 & 69   		 	 & 79				 	  &	76			 \\
Label Regularization & 79					 & 68    			 & 82				 	  &	76			 \\ \hline
\end{tabular}
\end{table}

\subsection{Image classification result}

In this section, we present image classification results using our proposed deep LLP model. For all experiments, we use the XCeption model \cite{DBLP:journals/corr/Chollet16a} with its training weights fit on ImageNet. The ImageNet dataset \cite{ILSVRC15} that is commonly used for image classification  has over a million images for 1,000 objects but does not have any class related to a human face or body. Because we have the best result with the maximum batch size 32 in the last section, we use it again for image classification task. To reduce memory consumption, we freeze first two blocks of XCeption network in the training phrase, and train the remaining layers.

To avoid overfitting and make the model robust to the wild condition of images, we apply various image distortions before each training iteration; we randomly rotate, flip (horizontally), shift (vertically and horizontally), shear, and zoom photos. Also, with probability of 80\%, we randomly crop the Viola-Jones detected face to avoid overfitting the background in images. We use Adam \cite{DBLP:journals/corr/KingmaB14}, an adaptive stochastic gradient descent algorithm, as an optimization algorithm, and we train all models for up to 20 epochs and report the accuracy on the validation set (Twitter annotated image profiles).

Table~\ref{tab:image:const} shows the weighted average F1 measure for different tasks using all constraint combinations. According to this table, the search constraint is more informative in all tasks, and using county with name constraints together has a poor result for race classification. Also, except for the {\bf race3} task, using all constraints together has the best result. By comparing this with text classification, because of search constraints, it is apparent that image classification has higher accuracy than text classification.

\begin{table}[t]
\centering
\caption{Comparison of F1 of image classification for different constraints on Twitter and CFD datasets:}
\label{tab:image:const}
\scalebox{0.85}{\begin{tabular}{lcccccc	}
Database           & TW          & CFD         & TW          & CFD         & TW          & CFD         \\
Task               & race2       & race2       & race3       & race3       & gender      & gender      \\ \hline
county             & 76          & 72          & 52          & 25          & 53          & 59          \\
name               & 61          & 30          & 54          & 25          & 88          & 65          \\
search             & 85          & 94          & 71          & 83          & 91          & 73          \\
county-name        & 61          & 31          & 52          & 25          & 93          & 68          \\
county-search      & 83          & 85          & \textbf{79} & 82          & 93          & 67          \\
name-search        & 81          & 91          & \textbf{79} & \textbf{85} & 92          & 68          \\
county-name-search & \textbf{92} & \textbf{98} & 77          & 75          & \textbf{95} & \textbf{75} \\ \hline
\end{tabular}}
\end{table}

This table also presents the F1 metric for CFD dataset. For {\bf race2}, the model has a very high result of 98\% F1 measure. This result reveals that race classification is much easier with high-quality frontal photos, as opposed to the noisier images from Twitter. On the other hand, while the gender classifier has very high F1 (95\%) for Twitter images, it has lower F1 for CFD dataset. We believe that is in part because the CFD images omit body features, and so the classifier must rely solely on face features.

\begin{figure}[ht!]
    \centering
    \subfloat[Effect of image distortion on race2]{\label{fig.dist}{\includegraphics[width=2.2in]{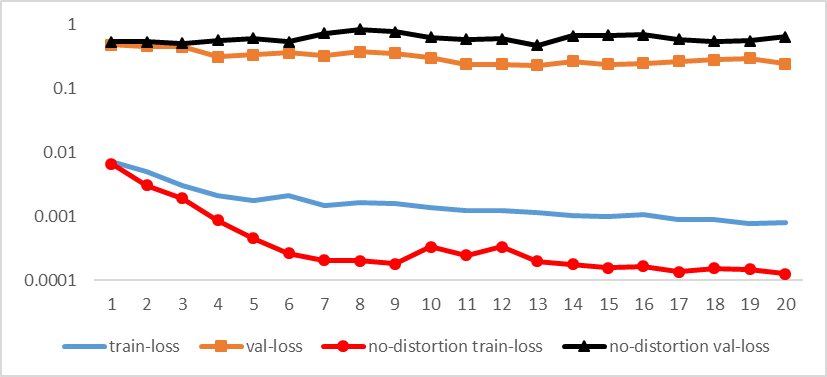} }}%
    \qquad
    \subfloat[Accuracy of tasks]{\label{fig.tasks}{\includegraphics[width=2.2in]{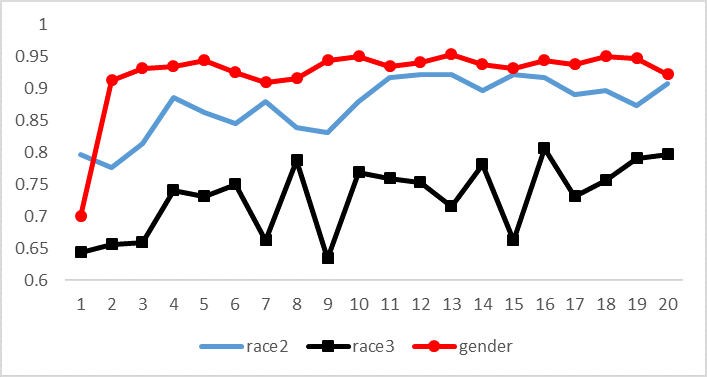} }}%
    \qquad
    \subfloat[Gender training loss]{\label{fig.train.loss}{\includegraphics[width=2.2in]{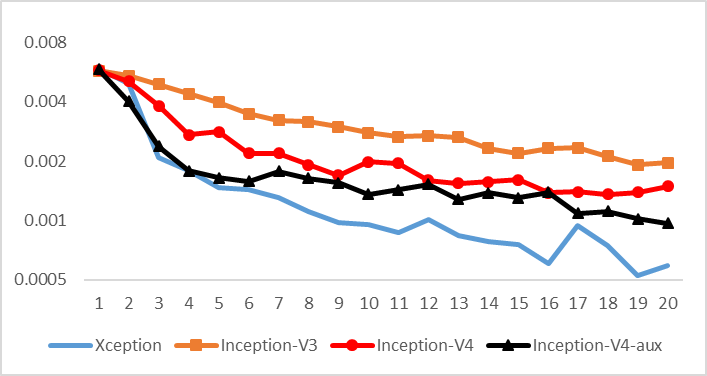} }}%
    \qquad
    \subfloat[Gender validation accuracy]{\label{fig.val.acc}{\includegraphics[width=2.2in]{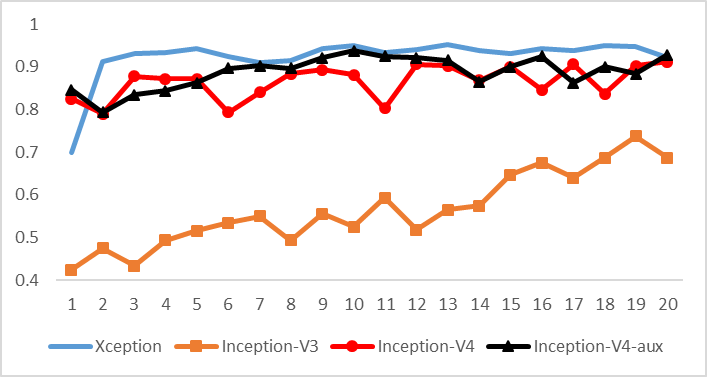} }}%
    \qquad
    \subfloat[Gender validation loss]{\label{fig.val.loss}{\includegraphics[width=2.2in]{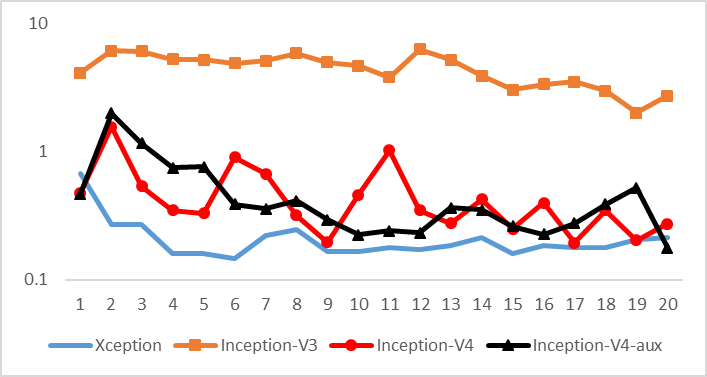} }}%
    \caption{Learning curve of different models.}%
\end{figure}

Since, on average, using all constraints has the highest average F1 measure on all tasks, in all next experiments we present the results of models that trains with all constraints. Figure~\ref{fig.tasks} illustrates the validation accuracy (on Twitter labeled data) of each training epoch for different classification tasks. Clearly, the gender classification has the highest accuracy and converges faster than other classes, and race3 has the lowest accuracy and converges slower.

To illustrate the impact of adding image distortion to make the model robust to the wild condition of pictures, Table~\ref{tab:image:networks} compares a model trained with random distortion using all constraints with the model without any distortion (Xception-no-distortion). According to this table, clearly, we need image manipulations for almost all tasks, and adding random distortion to photos has on average 7\% absolute improvement of the F1 measure. Similarly, Figure~\ref{fig.dist} shows the training and validation loss (KL-divergence) of race2 classification per epoch. According to this figure, the model that trains without image distortion overfits by converging to a lower training loss but with a higher validation loss.

To measure the effect of the underlying deep neural network, Table~\ref{tab:image:networks} demonstrates the average F1 of deep LLP models using various neural networks (with all constraints). We compare Xception with Inception-v3, Inception-v4, and Inception-v4-aux networks, and the former is same as Inception-v4, but has an auxiliary output layer as proposed by Szegedy (2016) \cite{DBLP:journals/corr/SzegedyIV16}. We add another {\bf Batch Averager} after the auxiliary layer and use KL-divergence for that too. 

According to this table, Inception-v3 has a poor result, and Inception-v4 does not converge for race3 task, but has the highest F1 measure (99\%) for race2 for CFD dataset. Also, the need of auxiliary layer for Inception-v4 is clear, and its improvement is significant. However, on average, in contrast to ImageNet reported results, Xception has a slightly better result than Inception-v4-aux. That maybe in part because the Inception-v4-aux is very deep and requires more data for the training phase.

\begin{table}[ht!]
	\centering
	\caption{Comparing underlying deep neural networks.}
	\label{tab:image:networks}
	\scalebox{.9}{\begin{tabular}{lcccccc}
	  \hline \hline
		Database  						 & TW    	  & CFD   		& TW    		& CFD   	 & TW     	& CFD    	 \\
		Task      						 & race2 	  & race2 		& race3 		& race3 	 & gender 	& gender 	 \\ [0.5ex] \hline
		Xception  						 & {\bf 92} & 98    		& {\bf 77}	& 75    	 & {\bf 95} & 75     	 \\
		Xception-no-distortion & 80    	  & 94    		& 71    		& 70    	 & {\bf 95}	& 63     	 \\
		Inception-v3  				 & 76			  & 98		 		& 68				&	67			 & 73			  & 50     	 \\
		Inception-v4 					 & 84			  & {\bf 99} &	53				& 25			 & 91			  & 62     	 \\
		Inception-v4-aux			 & 85			  & 96				&	70				& {\bf 84} & 94				& {\bf 81} \\ [1ex] \hline
	\end{tabular}}
\end{table}

Figures \ref{fig.train.loss}-\ref{fig.val.loss} illustrate training loss, validation accuracy, and validation loss of gender classification task using different models. According to these figures, Xception converges faster and has better results, and Inception-v3 has the worst results. Also, it is apparent that adding the auxiliary layer to Inception-v4 improves it, but it still has slightly lower result than Xception.

Finally, we compare deep LLP with four baselines. In our first two baselines, we compare deep LLP with supervised deep learning; we use Xception with its pre-trained weights, and train it for 50 epochs using labeled data. For {\bf Xception-wild} model, the training data is Twitter evaluation labeled data in wild conditions and we use CFD dataset with high quality images to train {\bf Xception-HQ} model. Our next two baselines are public face APIs. These APIs can detect multiple faces in the scene with multiple attributes. The {\bf Microsoft} face API\footnote{https://www.microsoft.com/cognitive-services/en-us/face-api} can identify gender but does not predict race. For race classification, we use {\bf Sightcorp} API\footnote{https://face.sightcorp.com/}, which is the only API supporting race recognition to the best of our knowledge, but it is still in the beta phase and does not have very accurate results. Table~\ref{tab:image:api} compares these baselines to our deep LLP model using all constraints. Our approach outperforms Xception-wild, Xception-HQ, and Sightcorp for all tasks and outperforms the Microsoft API in the wild condition. However, the latter has the better result for CFD dataset with clean, frontal images, which our method was not trained on.

\begin{table}[t]
	\centering
	\caption{Comparing deep LLP model (county-name-search) with baselines.}
	\label{tab:image:api}
	\begin{tabular}{lcccccc	}
		Database  		& TW    & CFD   & TW    & CFD   & TW     	 & CFD    	\\
		Task      		& race2 & race2 & race3 & race3 & gender 	 & gender 	\\ [0.5ex] \hline
		Deep LLP  		& 92    & 98    & 77    & 75    & 95 			 & 75     	\\
		Xception-wild	& ---   & 93    & ---   & 75    & ---      & 71     	\\
		Xception-HQ		& 85    & ---   & 71    & ---   & 82       & ---    	\\
		Microsoft 		& ---   & ---   & ---   & ---   & 80       & 90				\\
		Sightcorp 		& 24    & 53    & 21    & 46    & 22     	 & 64     	\\ [1ex] \hline
	\end{tabular}
\end{table}

\subsection{Co-training result}

In this section, we provide experimental results for Algorithm~\ref{alg:cot}. For these results, we initialize algorithm with the image view. (It is also possible to initialize it with the textual view, but we do not expect a significant difference.) In each iteration, the algorithm creates pseudo-bags, which are used in the next iteration (by switching the view). In this experiment, we use the same Twitter unlabeled data (with both image and text) as unlabeled samples. Thus, we use the same unlabeled data, but organize them into different bags with different label proportions for training. For text bags, we use county and name constraints, and for image bags, we use county, name, and search constraints.

Table~\ref{tab:cot:image} presents the result of the co-training algorithm for  image classification. In this table, the first column indicates the iteration of Algorithm~\ref{alg:cot}, and the first row (iteration zero) states the F1 measure of deep LLP model without co-training. This table only shows odd iterations (image classification steps) of the co-training algorithm. According to this table, the biggest improvement comes in the first iteration, which improves absolute F1 of race2 by 2\% (an error reduction of 25\%), and improves the absolute F1 of race3 by 4\% (an error reduction of 17\%).

\begin{table}[t]
\centering
\caption{The co-training results for image classification. Iteration 0 reports F1 measure before running the co-training algorithm.}
\label{tab:cot:image}
\begin{tabular}{lcccccc}
Database & TW    		& CFD   	 & TW    		& CFD   	 & TW     	& CFD      \\
Iteration& race2 		& race2 	 & race3 		& race3 	 & gender 	& gender   \\ \hline
0        & 92    		& {\bf 98} & 77    		& 75    	 & 95     	& 75       \\
1        & 94    		& {\bf 98} & 81    		& {\bf 87} & 95     	& 75     	 \\
3        & 94    		& {\bf 98} & {\bf 84} & 82    	 & 95     	& 77     	 \\ 
5        & {\bf 95} & {\bf 98} & 81    		& 86    	 & {\bf 96} & {\bf 80} \\ \hline
\end{tabular}
\end{table}

By applying the co-training algorithm, the most growth belongs to {\bf race3} with an absolute improvement of 7\% for Twitter and 12\% for CFD dataset. We believe that is in part because that the text classifier can detect Spanish words for Hispanic class and make better pseudo-bags for it.  For the {\bf race2} classification task, our co-training algorithm improves absolute F1 by 3\% for the Twitter dataset. The CFD dataset has already very high F1 measure (98\%), and does not have any growth by co-training steps.

\cut{To further illustrate training steps, Figure~\ref{fig.lc} shows the training loss (KL-divergence), the validation loss (KL-divergence), and the validation accuracy after each epoch for the 5\textsuperscript{th} iteration of co-training for image classification. According to this plot, the validation accuracy starts with 72.4\% and reaches to its highest in epoch 14, and then the model starts overfitting.

\begin{figure*}[t]
		\centering \includegraphics[scale=.4]{plots/lc.png}
    \caption{The learning curve: Shows training loss, validation loss, and validation accuracy for each epoch.}
    \label{fig.lc}
\end{figure*}
}

\begin{figure}[t]
		\centering \includegraphics[scale=0.15]{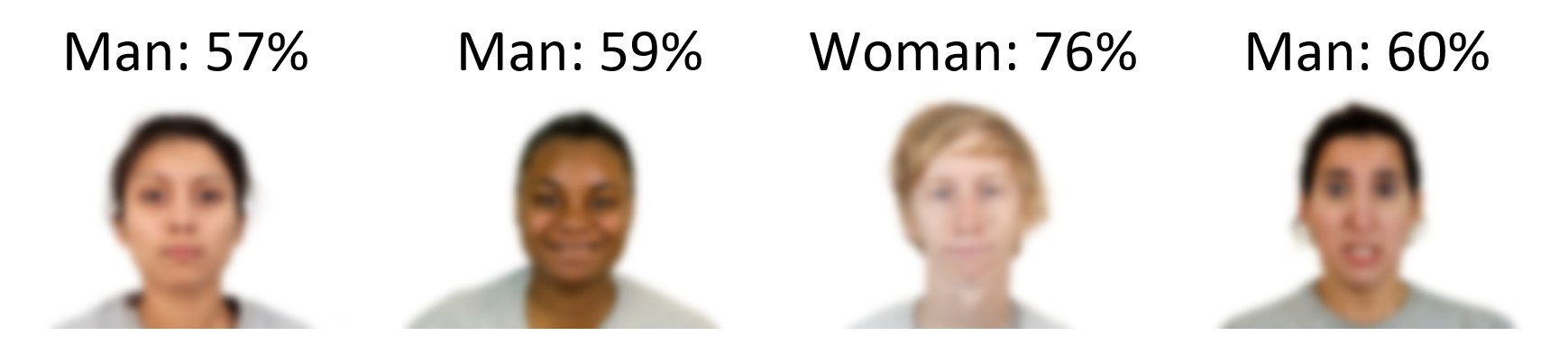}
    \caption{Examples misclassified by the image classification co-training model.}
    \label{fig.mc}
\end{figure}

Because the {\bf gender} classification already has a very high F1 measure (95\%) on Twitter, the co-training improves it by only 1\%. However, it increases the F1 on the CFD dataset by 5\%. We believe the lower accuracy for the {\bf gender} classification task of the CFD database is in part because of the lack of body features. In the absence of body features, the classifier often misclassifies a short hair woman as a man or a long hair man as a woman. Also, because CFD dataset has images with facial expression too, and our training data does not consider that, the classifier sometimes misclassifies a woman with an angry or a fearful expression as a man. The Figure~\ref{fig.mc} shows some misclassification samples with their predicted probability. (Images are blurred to protect privacy.)

Table~\ref{tab:cot:text} demonstrates the result of even iterations, text classification steps, of Algorithm~\ref{alg:cot}. Again, the first row shows the result of text classification without co-training. According to this table, the second iteration of co-training has the highest improvement (7\% on average across all tasks). The highest growth belongs to {\bf race2} classes, which improves by 10\% in absolute F1. This improvement is likely because the image classifier has a very high F1 for this task, and as a result, it creates accurate pseudo-bags for the text classifier.  Finally, the {\bf gender} classification grows only 3\%, since it has a little improvement from image classification.

\begin{table}[t]
\centering
\caption{The co-training results for text classification. Iteration 0 reports F1 measure before running the co-training algorithm.}
\label{tab:cot:text}
\begin{tabular}{ccccc}
Iteration & race2 	 & race3 		& gender 	 & Average  \\ \hline
0         & 80    	 & 73    	  & 83     	 & 79       \\
2         & 90   	   & 81   	  & {\bf 86} & 86       \\
4         & 91    	 & {\bf 83} & {\bf 86} & {\bf 87} \\
6         & {\bf 92} & {\bf 83} & 85     	 & {\bf 87} \\ \hline
\end{tabular}
\end{table}

In our final experiment, we select the last iteration for text classification (6\textsuperscript{th} iteration) and image classification (5\textsuperscript{th} iteration) of co-training algorithm. Then, we use them on the Twitter dataset and blend their predictions by using  soft voting. Table~\ref{tab:cot:ens} shows the result. According to this table, the ensemble improves image classification by average 2\%, and text classification by average 6\%. The ensemble produces the highest F1 for all tasks (except image classification for {\bf gender}, which has the same accuracy).

\begin{table}[ht!]
\centering
\caption{Results for the ensemble (soft voting) method, using the final iterations (text and image) of the co-training algorithm.}
\label{tab:cot:ens}
\begin{tabular}{lcccc}
Method    & race2 	 & race3 		& gender 	 & Average  \\ \hline
Text      & 92    	 & 83    		& 85     	 & 87       \\
Image     & 95    	 & 81    		& {\bf 96} & 91       \\
Ensemble  & {\bf 96} & {\bf 86} & {\bf 96} & {\bf 93} \\ \hline
\end{tabular}
\end{table}

\section{Conclusions and future work}
\label{sec.conclusions}

In this paper, we introduced two enhancements to deep learning methods for image and text classification: (1) a Batch Averager layer to enable LLP deep learning, and (2) a co-training method for combining deep LLP models trained on different views. We found that a deep model that is trained on population level data using proper constraints is comparable to  traditional supervised learning. This approach decreases the burden of human annotation and can readily be implemented with almost every publicly-available deep learning software packages.

We also found that for applications with two views (image and text) of features, a co-training algorithm leverages improvement of classification task for both views, and can be enhanced by an ensemble learning to achieve the highest precision.

In the future, we will investigate additional co-training algorithms for LLP, particularly investigating methods for improving pseudo-bag generation and applying it to a larger set of unlabeled data.


\section*{Acknowledgment}

This research was funded in part by the National Science Foundation under grants \#IIS-1526674 and \#IIS-1618244, and CCC Information Services Inc. provided a server (with multiple GPUs) to run deep learning models.




\bibliographystyle{IEEEtran}
\bibliography{IEEEabrv,face}

%



\end{document}